\documentclass[final]{cvpr}

\usepackage{times}
\usepackage{epsfig}
\usepackage{graphicx}
\usepackage{amsmath}
\usepackage{amssymb}
\usepackage{enumitem}

\usepackage{graphics}
\usepackage{floatrow}
\usepackage[label font=normalfont]{subfig}
\usepackage{multirow}
\usepackage{textpos}

\setlist{nosep} 

\usepackage[pagebackref=true,breaklinks=true,colorlinks,bookmarks=false]{hyperref}

\newcommand{\Section}[1]{\vspace{-1mm} \section{#1} \vspace{0mm}}
\newcommand{\SubSection}[1]{\vspace{0mm} \subsection{#1} \vspace{-1mm}}
\newcommand{\SubSubSection}[1]{\vspace{-1mm} \subsubsection{#1} \vspace{-1mm}}
\newcommand{\Paragraph}[1]{\vspace{1.25mm}\noindent\textbf{#1.}\hspace{0.5mm}}

\newcommand\Mark[1]{\textsuperscript#1}

\def\cvprPaperID{10612} 
\def\confYear{CVPR 2021}

\begin{document}
		
\begin{textblock*}{\textwidth}(0cm,0cm)
	\large\noindent{\copyright~2021 IEEE.  Personal use of this material is permitted.  Permission from IEEE must be obtained for all other uses, in any current or future media, including reprinting/republishing this material for advertising or promotional purposes, creating new collective works, for resale or redistribution to servers or lists, or reuse of any copyrighted component of this work in other works.}
\end{textblock*}
\thispagestyle{empty}

\title{Radar-Camera Pixel Depth Association for Depth Completion}

\author{
Yunfei Long\Mark{1}, Daniel Morris\Mark{1},  Xiaoming Liu\Mark{1}, \\
Marcos Castro\Mark{2},
Punarjay Chakravarty\Mark{2},
and Praveen Narayanan\Mark{2} \\ 
\Mark{1}Michigan State University, \Mark{2}Ford Motor Company \\
{\tt\small \{longyunf,dmorris,liuxm\}@msu.edu},
{\tt\small \{mgerard8,pchakra5,pnaray11\}@ford.com}
}

\twocolumn[{%
\renewcommand\twocolumn[1][]{#1}%
\maketitle
\pagenumbering{arabic}

\vspace{-1mm}
\noindent\begin{minipage}{\linewidth} 
 	\begin{center}
 	\captionsetup{font=small}
 	\begin{tabular}{@{}c@{}c@{}c@{}}
 	\includegraphics[trim = 190 27 210 30, clip, width=0.32\linewidth]{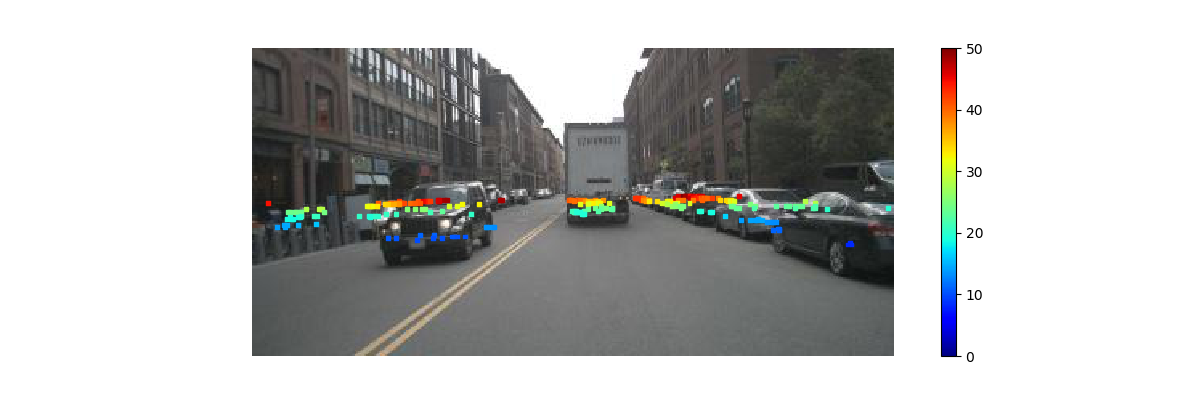} &
 	\includegraphics[trim = 190 27 210 30, clip, width=0.32\linewidth]{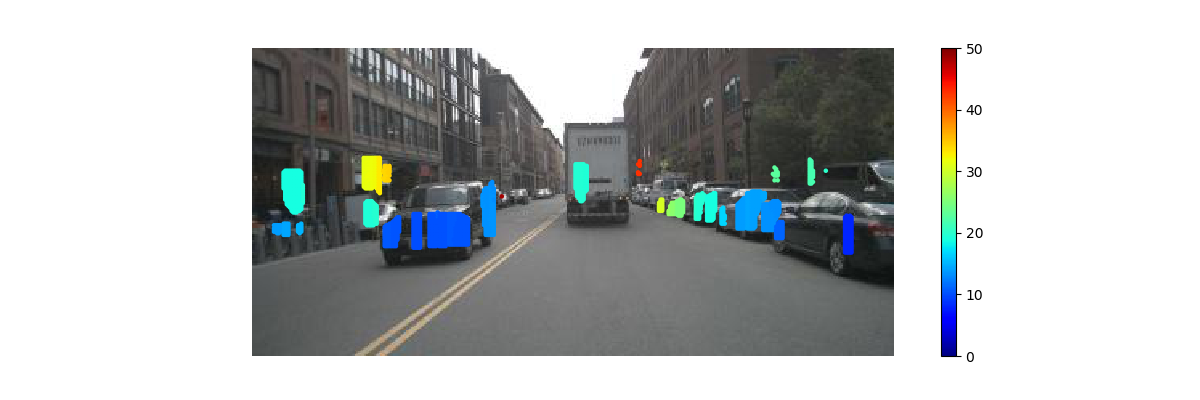} &
    \includegraphics[trim = 190 27 140 30, clip, width=0.37\linewidth]{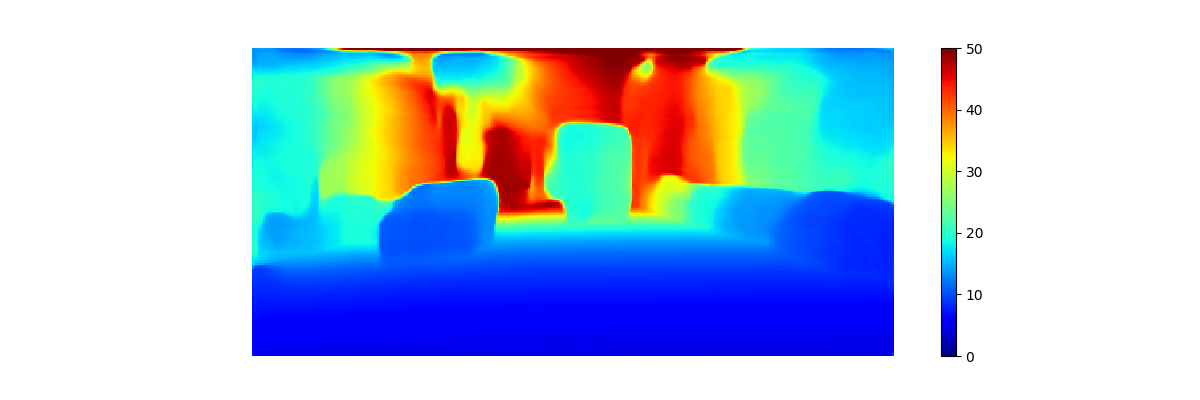} \vspace{-1mm}\\
    {\small (a)} &  {\small (b)}  &  {\small (c)} \\
    \end{tabular}
	\vspace{-3mm}
	\captionof{figure}{\small Radar-camera depth completion: (a) an image with $0.3$ seconds ($5$ sweeps) of radar hits projected onto it, (b) enhanced radar depths at confidence level $0.9$ eliminate occluded pixels and expand visible hits, and (c) final predicted depth through depth completion. }
	\label{fig:radar_augmentation}
	\end{center}  \vspace{4mm}
\end{minipage}
}]

\begin{abstract}
While radar and video data can be readily fused at the detection level, fusing them at the pixel level is potentially more beneficial.
This is also more challenging in part due to the sparsity of radar, but also because automotive radar beams are much wider than a typical pixel combined with a large baseline between camera and radar, which results in poor association between radar pixels and color pixel.  
A consequence is that depth completion methods designed for LiDAR and video fare poorly for radar and video.  Here we propose a radar-to-pixel association stage which learns a mapping from radar returns to pixels.  
This mapping also serves to densify radar returns.  Using this as a first stage, followed by a more traditional depth completion method, we are able to achieve image-guided depth completion with radar and video.  We demonstrate performance superior to camera and radar alone on the nuScenes dataset.
Our source code is available at \url{https://github.com/longyunf/rc-pda}.

\end{abstract}

\vspace{-2mm}

\section{Introduction}

We seek to incorporate automotive radar as a contributing sensor to $3$D scene estimation. 
While recent work fuses radar with video for the objective of achieving improved object detection~\cite{Chadwick:DistantVehicle:2019,Nabati:RadarRegionProposal:2019,lim:2019:radar,Meyer:2019:radar,Nobis:2019:radar}, here we aim for pixel-level fusion of depth estimates, and ask if {\it fusing  video with radar can lead to improved dense depth estimation of a scene}.  

Up to the present, outdoor depth estimation has been dominated by LiDAR, stereo, and monocular techniques.  The fusion of LiDAR and video has lead to increasingly accurate dense depth completion~\cite{Imran2019}.  
At the same time, radar has been relegated to the task of object detection in vehicle's Advanced Driver Assistance Systems (ADAS)~\cite{Marti:ADAS:2019}.  However, phased array automotive radar technologies have been advancing in accuracy and discrimination~\cite{Hasch:AutomotiveRadar:2015}.  Here we investigate the suitability of using radar instead of LiDAR for the task of dense depth estimation.  Unlike LiDAR, automotive radars are already ubiquitous, being integrated in most vehicles for collision warning and similar tasks.  If successfully fused with video, radar could provide an inexpensive alternative to LiDARs for $3$D scene modeling and perception.   
However, to achieve this, attentive algorithm design is required in order to overcome some of the limitations of radar, including coarser, lower resolution, and sparser depth measurements than typical LiDARs.

This paper proposes a method to fuse radar returns with image data and achieve depth completion; namely a dense depth map over pixels in a camera.  We develop a two-stage algorithm.  The first stage builds an association between radar returns and image pixels, during which we resolve some of the uncertainty in projecting radar returns into a camera. 
In addition, this stage is able to filter occluded radar returns and ``densify'' the projected radar depth map along with a confidence measure for these associations
(see Fig.~\ref{fig:radar_augmentation} (a,b)).  Once a faithful association between radar hits and camera pixels is achieved, the second stage uses a more standard depth completion approach to combine radar and image data and estimate a dense depth map, as in Fig.~\ref{fig:radar_augmentation}(c).

A practical challenge to our fusion goal is the lack of public datasets with radar.  
KITTI~\cite{Geiger2012CVPR}, the dataset used most extensively for LiDAR depth completion, does not include radar and nor do the Waymo~\cite{sun2020scalability} or ArgoVerse~\cite{chang2019argoverse} datasets.  
The main exception is nuScenes~\cite{nuscenes2019} and the small Astyx~\cite{Meyer:RadarDataset:2019} which have radar, but unfortunately do not include a dense, pixel-aligned depth map as created by Uhrig~\emph{et al.}~\cite{Uhrig2017THREEDV}. Similarly, the Oxford Radar Robot Car dataset~\cite{RadarRobotCarDatasetICRA2020} includes camera, LiDAR and raw radar data, but no annotations are available for scene understanding. 
As a result, all experiments of this work will use the nuScenes dataset along with its annotations.  However, we find single LiDAR scans insufficient to train depth completion, and so accumulate scans to build semi-dense depth maps for training and evaluating depth completion.

The main contributions of this work include:
\begin{itemize}
 \item Radar-camera pixel depth association that upgrades the projection of radar onto images and prepares a densified depth layer.
 \item Enhanced radar depth that improves radar-camera depth completion over raw radar depth.
 \item LiDAR ground truth accumulation that leverages optical flow for occluded pixel elimination, leading to higher quality dense depth images.
\end{itemize} 

\section{Related Work}

\Paragraph{Radar for ADAS}
Frequency Modulated Continuous Wave radars are inexpensive and all-weather, and have served as the key sensor for modern ADAS.  Ongoing advances are improving radar resolution and target discrimination~\cite{Hasch:AutomotiveRadar:2015}, while convolutional networks has been used to add discriminative power to radar data, moving beyond target detection and tracking to include classifying road environments~\cite{Lee:RadarRoadRecognition:2019,Sim:radar-for-road-environments:2020}, and seeing beyond-line-of-sight targets~\cite{Scheiner:DopplerSeeingAroundCorners:2020}.  Nevertheless, the low spatial resolution of radar means that the $3$D environment, including object shape and classification, are only coarsely obtained.  A key path to upgrading the capabilities of radar is through integration with additional sensor modalities~\cite{Marti:ADAS:2019}.

\Paragraph{Radar-camera fusion}
Early fusion of video with radar, such as~\cite{ji:radar:2008}, relied on radar for cueing image regions for object detection or road boundary estimation~\cite{Janda:RoadBoundary:2013}, or used optical flow to improve radar tracks~\cite{Garcia:RadarFlowTracks:2012}.  
With the advent of deep learning, much more extensive multi-modal fusion has become possible~\cite{Feng:MultiModealFusion:2020}.
However, to the best of our knowledge, no prior work has conducted pixel-level dense depth fusion between radar and video.

\Paragraph{Radar-camera object detection}
Object detection is a key task in $3$D perception~\cite{kinematic-3d-object-detection-in-monocular-video}.
There has been significant recent interest in combining radar with video for improved object detection. In~\cite{Chadwick:DistantVehicle:2019},  ResNet blocks~\cite{He:2016:DeepResNet} are used to combine both color images and image-projected radar returns to improve longer-range vehicle detection.  
In~\cite{lim:2019:radar}, an FFT applied to raw radar data generates a polar detection array which is merged with a bird's eye projection of the camera image, and targets are estimated with a single shot detector~\cite{liu:2016:ssd}.  In~\cite{Meyer:2019:radar}, features from both images and a bird's-eye representation of radar enter a region proposal network that outputs bounding boxes~\cite{Simonyan:VGG:2015}. An alternative model for radar hits is a $3$m vertical line on the ground plane which is projected into the image plane by~\cite{Nobis:2019:radar}, and combined with VGG blocks to classify vehicle detections at multiple scales.  
Our work differs fundamentally from these methods in that our goal is dense depth estimation, rather than object classification.  But we do share similarity in radar representation: we project radar hits into an image plane.  However, the key novelty in our work is that we learn a neighborhood pixel association model for radar hits, rather than relying on projected circles~\cite{Chadwick:DistantVehicle:2019} or lines~\cite{Nobis:2019:radar}.  

\Paragraph{LiDAR-camera depth completion} 
Our task of depth estimation has the same goal as LiDAR-camera depth completion~\cite{jaritz2018sparse,qiu2019deeplidar,xu2019depth,Imran2019,depth-completion-with-twin-surface-extrapolation-at-occlusion-boundaries}.  However, radar is far sparser than LiDAR and has lower accuracy, which makes these methods unsuitable for this task. 
Our radar enhancement stage densifies the projected radar depths, followed by a more traditional depth completion architecture.

\Paragraph{Monocular depth estimation}
Monocular depth inference may be supervised by LiDAR or self-supervised.
Self-supervised methods learn depth by minimizing photometric error between images captured by cameras with known relative positions. 
Additional constrains such as semantics segmentation~\cite{zhu2020edge, ramirez2018geometry}, optical flow~\cite{janai2018unsupervised, vijayanarasimhan2017sfm}, surface normal~\cite{yang2017unsupervised} and proxy disparity labels~\cite{watson2019self} improve performance.  Recently, self-supervised PackNet~\cite{Guizilini:packnet:2020} has achieved competitive results.   Supervised methods~\cite{fu2018deep, diaz2019soft} include continuous depth regression and discrete depth classification. BTS~\cite{lee2019big} achieves state of the art by improving upsampling  via additional plane constraints, and more recently~\cite{ren2020suw} combines supervised and self-supervised methods.
Our goal is not monocular depth estimation, but rather to improve what is achievable from monocular depth estimation through fusion with radar.

\begin{figure*}[t!]
\captionsetup{font=small}
    \centering
    \includegraphics[width=\textwidth]{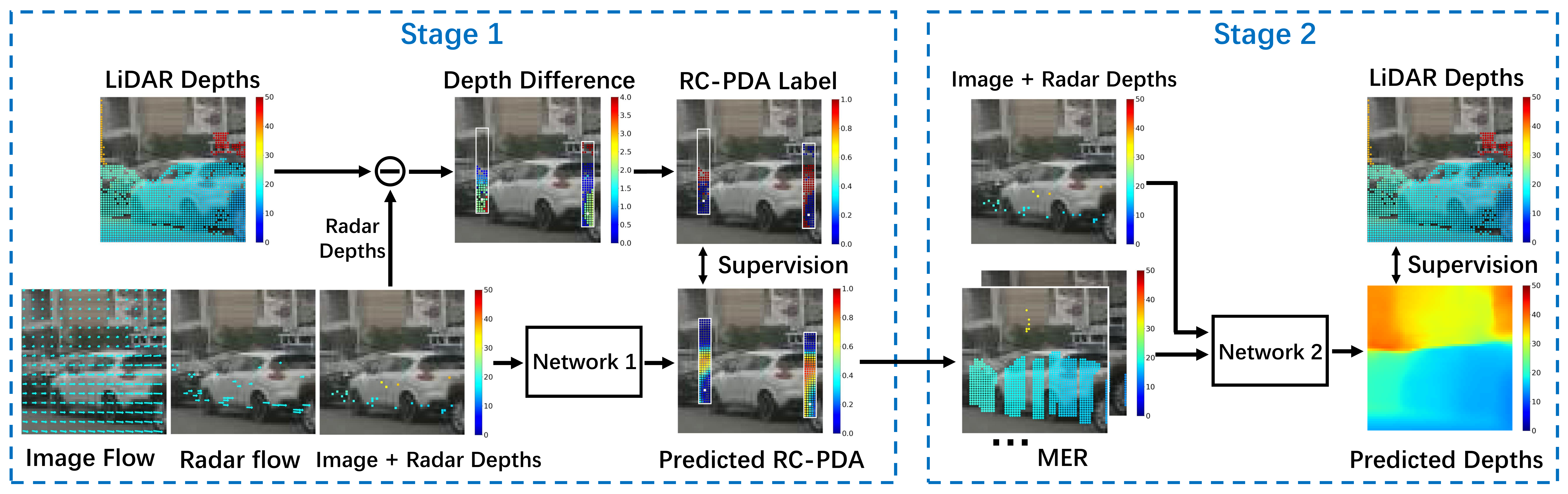}
    \vspace{-6mm}
     \caption{\small Our two-stage architecture.  Network $1$ learns $N$-channel \emph{radar-camera pixel depth association} (RC-PDA), here illustrated for two radar pixels (marked with white squares) on their neighboring pixels (white boxes). The RC-PDA is converted into a \emph{multi-channel enhanced radar} (MER), and input to Network $2$ which performs image-guided depth completion.  \vspace{-3mm}}
    \label{fig:network_arch}
\end{figure*}

\Section{Method}

While there are a variety of data-spaces in which radar can be fused with video, the most natural, given our objective of estimating a high resolution depth map, is in the image space.  But this immediately presents a problem: to which pixel in an image does a {\it radar pixel} belong?  By radar pixel we mean a simple point projection of the estimated $3$D radar hit into the camera. The nuScenes dataset~\cite{nuscenes2019} provides extrinsic and intrinsic calibration parameters needed to map the radar point clouds from the radar coordinates system to the egocentric and camera coordinate systems.

Assuming that the actual depth of the image pixel is the same as the radar pixel depth turns out to be fairly inaccurate. We describe some of the problems with this model, and then propose a new pixel association model.  We present a method for building this new model and show its benefit by incorporating it into radar-camera depth completion. Fig.~\ref{fig:network_arch} shows the diagram of the proposed method.

\SubSection{Radar Hit Projection Model}

\begin{figure}[t!]
\captionsetup{font=small}
    \centering
    \includegraphics[trim=6 0 0 0,clip,width=\linewidth]{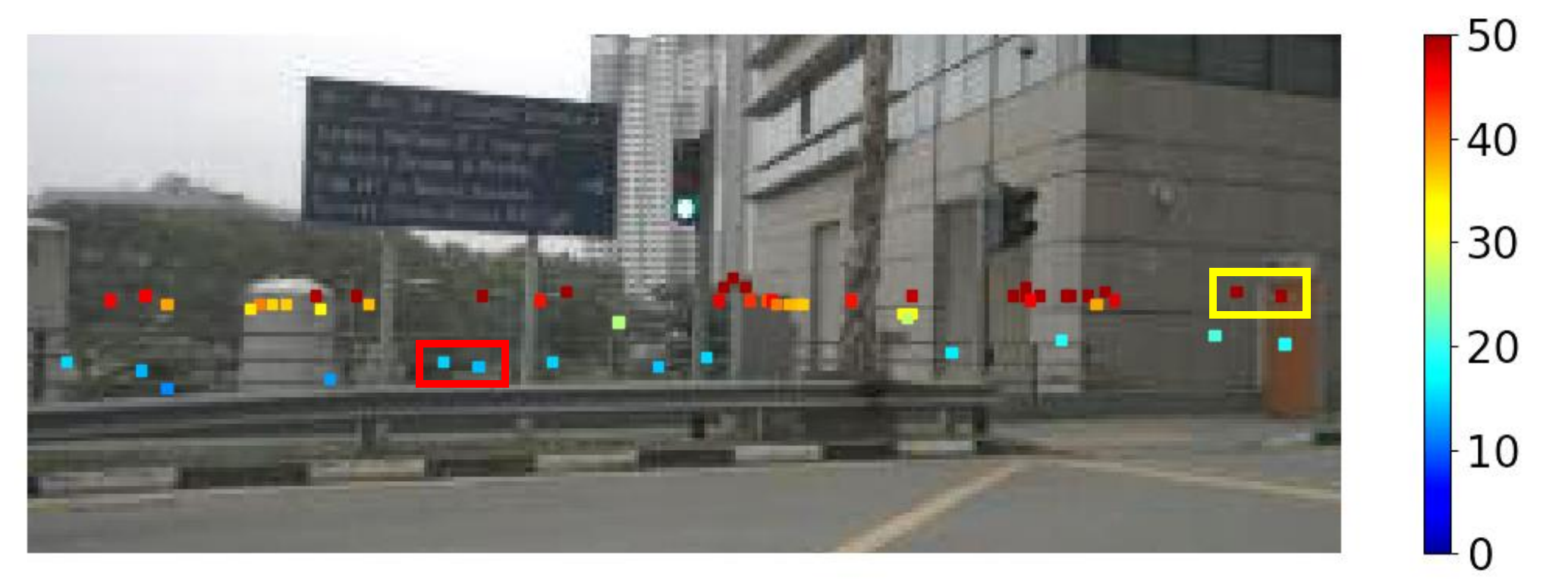}
    \vspace{-7mm}
    \caption{\small Examples of radar hits projected into a camera.  While the hits project into the vicinity of the target that they hit, their image position can be quite different from their actual location. For example, radar depths in the yellow/red box are larger/smaller than corresponding image depths (meters).}
    \label{fig:radar_examples}
\end{figure}

Single-row scanning automotive radars can be modeled as measuring points in a plane extending usually horizontally (relative to the vehicle platform) in front of the vehicle, as in~\cite{Chadwick:DistantVehicle:2019}. While radars can measure accurate depth, often the depth they give when projected into a camera is incorrect, as can be seen in the examples in Fig.~\ref{fig:radar_examples}.  An important source of this error is the large width of radar beams which means that the hits extend well beyond the assumed horizontal plane. In other words, the height of measured radar hits is inaccurate~\cite{nabati2021centerfusion}.

In addition to beam width, another source of projected point depth difference is occlusion caused by the significant baseline between radars on the grill, and cameras on the roof or driver mirror. Further, these depth differences only increase when radar hits are accumulated over a short interval, and thus more opportunity for occlusions.

In addition to pixel association errors, we are faced with the problem that automotive radars generate far sparser depth scans than LiDAR.  There is typically a {\it single row} of returns, rather than anywhere up to $128$ rows in LiDAR, and the azimuth spacing of radar returns can be an order of magnitude greater than LiDAR.  This sparsity significantly increases the difficulty in depth completion.  One solution is to accumulate radar pixels over a short time interval, and to account for their $3$D position using both ego-motion and radial velocity.  Nevertheless, this accumulation introduces additional pixel association errors (in part from not having tangential velocity) and more opportunities for occlusions.

\begin{figure*}[t!]
\captionsetup{font=small}
    \centering
    \begin{tabular}{@{}c@{}c@{}}
    \includegraphics[trim=196 245 239 145,clip,width=0.54\linewidth]{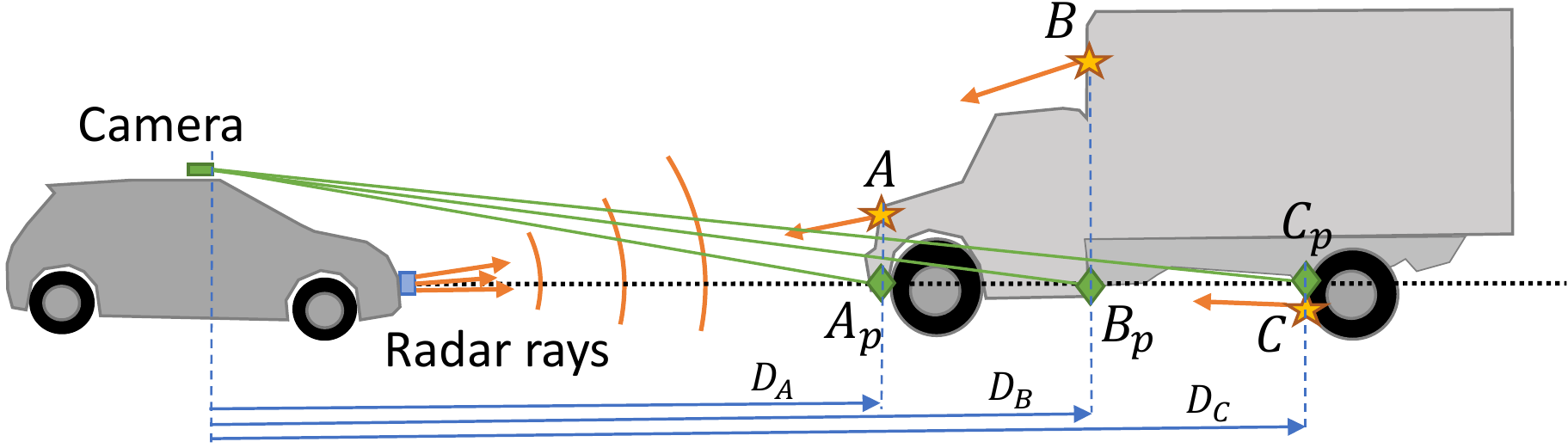} &
    \includegraphics[trim=160 300 270 50,clip,width=0.45\linewidth]{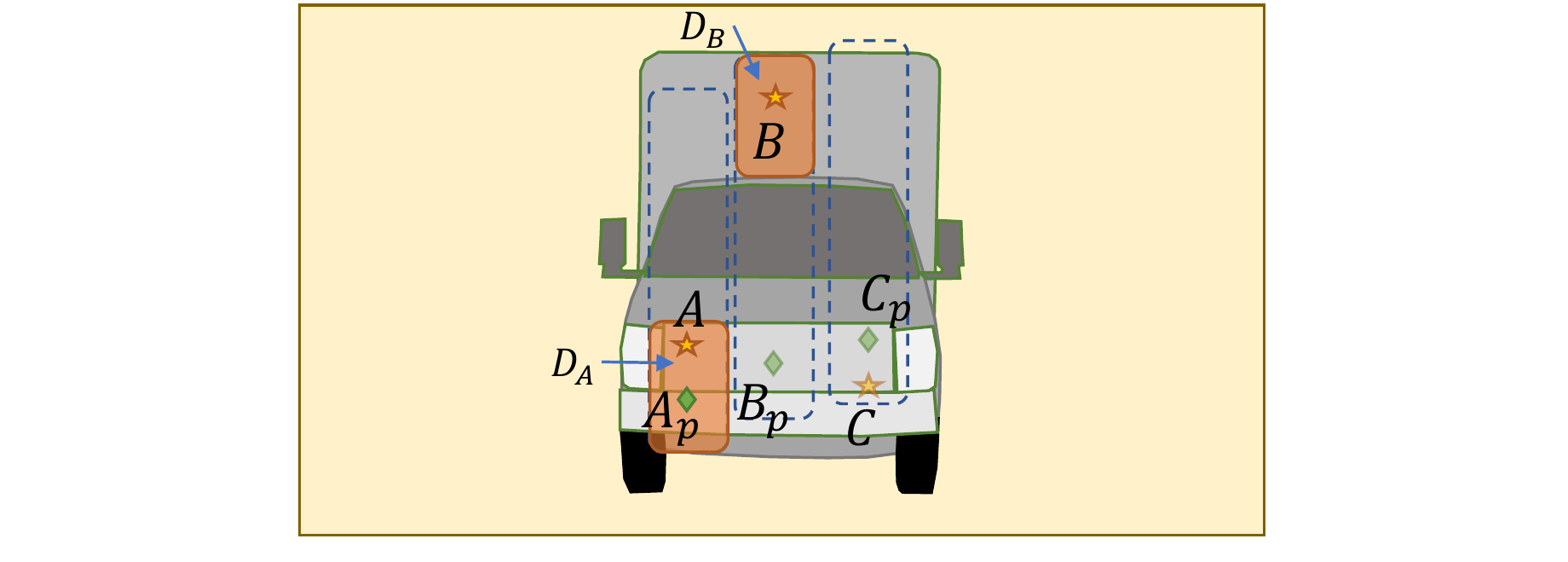} \vspace{-1mm}\\
     {\small (a)} &  {\small (b)} \vspace{-3mm}\\
    \end{tabular}    
    \caption{\small Illustration of depth differences between camera and radar, and how our proposed association method (Pixel Depth Association: RC-PDA) can address this. (a) Radar hits are modeled on a ground-parallel plane (dashed black line).  Actual returns may be outside this plane, as illustrated with orange stars $A,B,C$ at depths $D_A,D_B,D_C$ respectively.  We project the corresponding in-plane points, $A_p,B_p,C_p$ (green diamonds), into the camera, and call these the \emph{radar pixels}. (b) The camera view showing the radar pixels $A_p,B_p,C_p$.  Now the true image depth of these pixels is $D_A$, the front of the truck, which agrees only with $A_p$ which is visible, and not for $B_p$ and $C_p$ which are occluded.  This illustrates why radar pixel depths are often incorrect from the camera perspective.  Finding associations from radar pixels to the projected true points $A,B,C$ would solve this, but is difficult.  Rather, we seek a neighborhood \emph{depth association} for each radar pixel, that specifies which pixels within a neighborhood (dashed blue regions) have the same depth as the radar pixel, shown here by the orange regions.  For example, the orange pixels in the neighborhood of $A_p$ have a RC-PDA of $1$ while the remaining neighborhood pixels have a RC-PDA of $0$, all relative to $A_p$. See Sec.~\ref{sec:RCPDA} for details.  }
    \label{fig:radar_projection_model}
\end{figure*}

\SubSection{Radar-Camera Pixel Depth Association}
\label{sec:RCPDA}
In using radar to aid depth estimation we face the problem of determining which, if any, point in the image does a radar return correspond to?  This radar pixel to camera pixel association is a difficult problem, and we do not have ground truth to determine this. Thus we reformulate this problem slightly to make it more tractable.

The new question we ask is: ``Which pixels in the vicinity of the projected radar pixel have the same depth of that radar return?"  We call this \emph{Radar-Camera Pixel Depth Association}: RC-PDA or simply PDA.  
It is a one-to-many mapping, rather than one-to-one mapping, and has four key advantages.   
First, we do not need to distinguish between many good but ambiguous matches and rather can return many pixels with the same depth.  
This simplifies the problem. 
Second, by associating the radar return with multiple pixels, our method explicitly densifies the radar depth map, which facilitates the second stage of full-image depth estimation.  
Third, our question simultaneously addresses the occlusion problem; if there are no nearby pixels with that depth, then the radar pixel is automatically inferred to be occluded. 
Fourth, we are able to leverage a LiDAR-based ground-truth depth map as the supervision, rather than a difficult-to-define ``ground truth" pixel association.
Fig.~\ref{fig:radar_projection_model} illustrates image depths obtained from raw radar projections and RC-PDA around each radar pixel. It shows the height errors of measured radar points and how some hits visible to the radar are occluded from the camera.

\SubSubSection{RC-PDA Model}

\begin{figure}[t!]
\captionsetup{font=small}
    \centering
    \begin{tabular}{@{}c@{}c@{}c@{}}
	\includegraphics[trim=140 320 640 80,clip,width=0.32\linewidth]{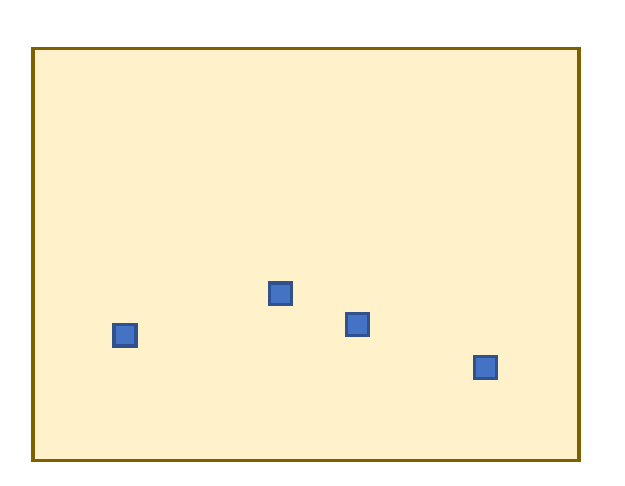} &
	\includegraphics[trim=140 320 640 80,clip,width=0.32\linewidth]{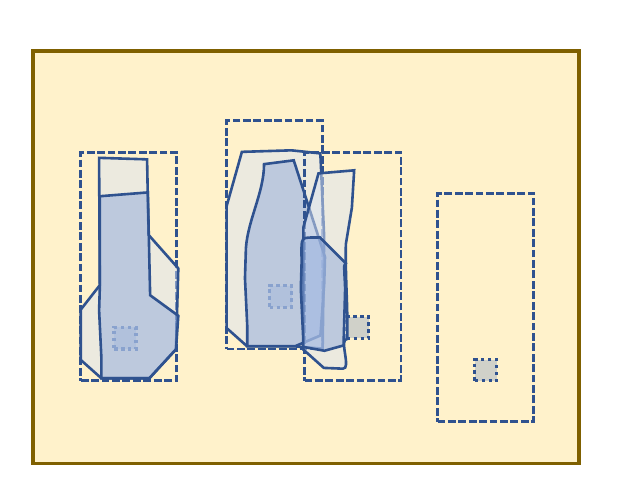} &
	\includegraphics[trim=140 320 640 80,clip,width=0.32\linewidth]{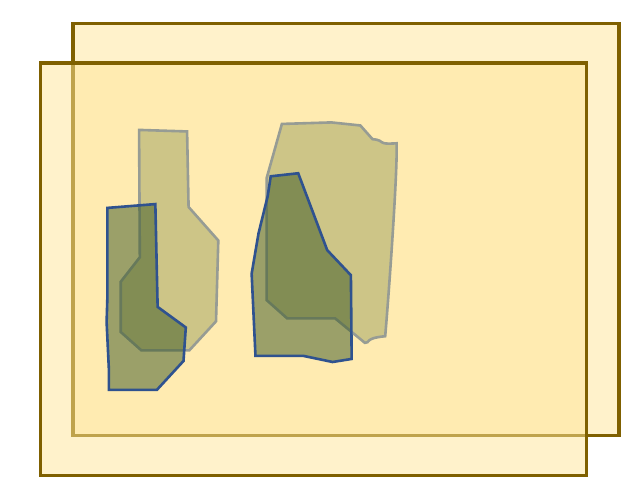} \vspace{-1mm}\\
     {\small (a)} &  {\small (b)} &  {\small (c)} \vspace{-2mm}\\
    \end{tabular}
\caption{\small Overview of our radar depth representation. (a) Radar pixels indicate sparse depth in image space. (b) For each radar pixel, a Pixel Depth Association (RC-PDA) probability over neighboring pixels is calculated, indicated with shaded contours.  (c) Radar pixel depths are propagated to neighboring pixels to create a  Multi-channel Enhanced Radar (MER) image.  Each channel is a densified depth at a given confidence level. \vspace{-4mm}}
    \label{fig:PDA_MER}
\end{figure}

We model RC-PDA over a neighborhood around the projected radar pixel in the color image.  
At each radar pixel we define a patch around the radar location and seek to classify each pixel in this patch as having the same depth or not as the radar pixel, within a predetermined threshold. A similar connectivity model has been used for image segmentation~\cite{kampffmeyer2018connnet}.
Radar pixels and the patches around them are illustrated in Figs.~\ref{fig:PDA_MER}(a) and (b), respectively.

The connection to each pixel in a $h\times w$ neighborhood has $N=w h$ elements, and can be encoded as an $N$-channel RC-PDA which we label $\mathbf{A}(i,j,k)$, where $k=1,\cdots,N$.  
Here $(i,j)$ is the radar pixel coordinate, and the $k$'th neighbor has offset $(i_k,j_k)$ from $(i,j)$.  Now the label for $\mathbf{A}(i,j,k)$ is 1 if the neighboring pixel has the same depth as radar pixel and 0 otherwise.  More precisely, if $E_{ijk}=d(i,j)-d_T(i+i_k,j+j_k)$ is the difference between radar pixel depth, $d(i,j)$, and the neighboring LiDAR pixel depth, $d_T(i+i_k,j+j_k)$, and $\tilde{E}_{ijk}=E_{ijk}/d(i,j)$ is the relative depth difference, then:
\begin{equation}
\label{eq:aff_label}
\mathbf{A}(i,j,k) =\left\{
\begin{aligned}
1 & , & \text{if } (|E_{ijk}| < T_a) \wedge (|\tilde{E}_{ijk}| < T_r)  \\
0 & , & \text{otherwise}.
\end{aligned}
\right.
\end{equation}
We note that labels $\mathbf{A}(i,j,k)$ are only defined when there is both a radar pixel at $(i,j)$ and a LiDAR depth $d_T(i+i_k,j+j_k)$. We define a binary weight $w(i,j,k)\in\{0,1\}$ to be $1$ when both conditions are satisfied and $0$ otherwise.  During training we minimize the weighted binary cross entropy loss~\cite{kampffmeyer2018connnet} between labels $\mathbf{A}(i,j,k)$ and predicted RC-PDA:
\begin{equation}
\begin{split}
    \mathcal{L}_{CE} =  \sum_{i,j,k} w(i,j,k)[&-\mathbf{A}(i,j,k) z(i,j,k) \\
                                              & + \log(1 + \exp(z(i,j,k)))].
\end{split}
\label{eq:sce_loss}
\end{equation}
The network output, $z(i,j,k)$, is passed through a Sigmoid to obtain $\hat{\mathbf{A}}(i,j,k)$, the estimated RC-PDA.

Our network thus predicts a RC-PDA confidence in a range of $0$ to $1$ representing the probability that each pixel in this patch has the same depth as the radar pixel.  This prediction also applies to the image pixel at the same coordinates as the radar pixel, $\it{i.e.}$ $(i,j)$, as like other pixels, the depth at this image pixel may differ from the radar depth for a variety of reasons, including those illustrated in Fig.~\ref{fig:radar_projection_model}.

\SubSection{From RC-PDA to MER}
\label{sec:pda_mer}
The RC-PDA gives the probability that neighboring pixels have the same depth as the measured radar pixel.  We can convert the radar depths along with predicted RC-PDA into a partially filled depth image plus a corresponding confidence as follows.  
Each of $N$ neighbors to a given radar pixel is given depth $d(i,j)$ and confidence $\hat{\mathbf{A}}(i,j,k)$.  
If more than one radar depth is expanded to the same pixel, the radar depth with the maximum RC-PDA is kept. 
The expanded depth is represented as $\mathbf{D}(i,j)$ with confidence $\hat{\mathbf{A}}(i,j)$.  
Now many of the low-confidence pixels will have incorrect depth.  
Instead of eliminating low-confidence depths, we convert this expanded depth image into a multi-channel image where each channel $l$ is given depth $\mathbf{D}(i,j)$ if its confidence $\hat{\mathbf{A}}(i,j)$ is greater than a channel threshold $T_l$, where $l=1,\cdots,N_e$ and $N_e$ is the total number of channels of the enhanced depth.
The result is a \emph{Multi-Channel Enhanced Radar} (MER) image with each channel representing radar-derived depth at a particular confidence level (see Fig.~\ref{fig:PDA_MER}(c)).  

Our MER representation for depth can correctly encode many complex cases of radar-camera projection, a few of which are illustrated in Fig.~\ref{fig:radar_projection_model}.  These cases include when radar hits are occluded and no nearby pixels have similar depth.  They also include cases where the radar pixel is just inside or just outside the boundary of a target.  In each case, those nearby pixels with the same depth as the radar can be given the radar depth with high confidence, while the remaining neighborhood pixels are given low confidence, and their depth are specified on separate channels of the MER.

The purpose of using multiple channels for depths with different confidences in MER is to facilitate the task of Network $2$ in Fig.~\ref{fig:network_arch} in performing the dense depth completion.  High confidence channels give the greatest benefit, but low confidence channels may also provide useful data.  In all cases they densify the depth beyond single radar pixels, easing the depth completion task.

\SubSection{Estimating RC-PDA}

We next select inputs to Network $1$ in Fig.~\ref{fig:network_arch} from which it can learn to infer the RC-PDA. 
These are the image, the radar pixels with their depths as well as the image flow and the radar flow from current to a neighboring frame.  Here we briefly explain the intuition for each of these.  

The image provides scene context for each radar pixel, as well as object boundary information.  The radar pixels provide depth for interpreting the context and a basis for predicting the depth of nearby pixels.  As radar is very sparse, we accumulate radar from a short time history, $0.3$ seconds, and transform it into the current frame using both ego-motion and the radial velocity similar to that done in~\cite{Nobis:2019:radar}.

\begin{figure}[t!]
\captionsetup{font=small}
	\centering
	\scalebox{1.1}{
    \begin{tabular}{@{}c@{}c@{}c@{}}
	\includegraphics[width=1.17in]{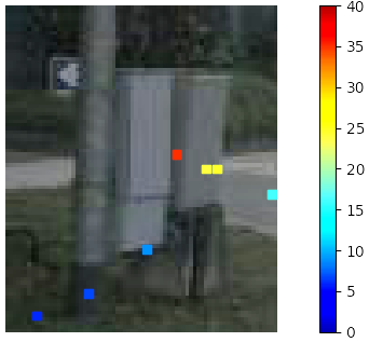} &
	\includegraphics[width=0.89in]{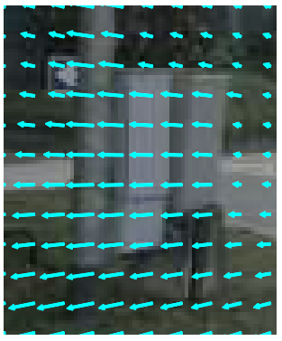} &
	\includegraphics[width=0.88in]{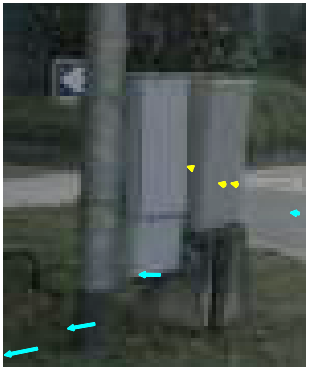}  \vspace{-1mm}\\
     \footnotesize{  (a) Radar depth} &  \footnotesize{  (b) Optical flow} & \footnotesize{  (c) Radar flow} \vspace{-2mm}\\
    \end{tabular}   }
	\caption{\small An example of how radar scene flow and optical flow differences are used to infer occlusions of radar pixels. The radar flows are plotted as yellow if the $L_2$ norm of radar/optical flow differences are larger than a threshold.  Note that we do not explicitly filter radar, rather provide flow to Network $1$ in Fig.~\ref{fig:network_arch} so that it can implicitly filter radar while estimating RC-PDA. \vspace{-3mm}}
	\label{Figure:flows} 
\end{figure}

Now a pairing of image optical flow and radar scene flow provides an occlusion and depth difference cue.  For static objects, the optical flow should exactly equal radar scene flow, when the pixel depth is the same as radar pixel depth.  Conversely, radar pixels that are occluded from the camera view will have different scene flow from the optical flow of a static object occluding them (Fig.~\ref{Figure:flows}). Similarly, objects moving radially will have consistent flow.  By providing flow, we expect that Network $1$ will learn to leverage flow similarity in predicting RC-PDA for each radar pixel.

\begin{figure}[t!]
\captionsetup{font=small}
	\centering
    \begin{tabular}{@{}c@{}c@{}c@{}}
	\includegraphics[trim=0 0 0 0,clip,width=2.8cm]{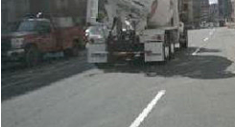} &
	\includegraphics[trim=0 0 0 0,clip,width=2.8cm]{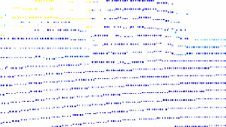} &
	\includegraphics[trim=0 0 0 0,clip,width=2.8cm]{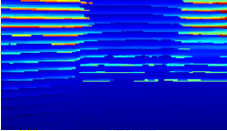}  \vspace{-1mm}\\
     \footnotesize{ (a) } &  \footnotesize{ (b)} & \footnotesize{ (c) } \vspace{-3mm}\\
    \end{tabular} 
	\caption{\small We noticed that when LiDAR with a regular scan pattern, as in (b) for image (a), is used to train depth completion, our network learns to predict the LiDAR points well, but not the remaining pixels.  This leaves large artifacts, as in (c), and motivates us to create a semi-dense depth LiDAR training set. \vspace{-2mm}} 
	\label{fig:artifacts}
\end{figure}

\SubSection{LiDAR-based Supervision}
\label{sec:lidar-supervision}
To train both the RC-PDA and the final dense depth estimate, we use a dense ground truth depth.  This is because, as illustrated in Fig.~\ref{fig:artifacts}, training with sparse LiDAR leads to significant artifacts.   We now describe how we build a semi-dense depth image from LiDAR scans.

\SubSubSection{LiDAR Accumulation}

To our knowledge, there is no existing public dataset specially designed for depth completion with radar. 
Thus we create a semi-dense ground truth depth from nuScenes dataset, a public dataset with radar data and designed for object detection and segmentation. 
We use the $32$-ray LiDAR as depth label and notice that the sparse depth label generated from a single frame will lead to a biased model predicting depth with artifacts, {\it i.e.}, only predictions for pixels with ground truth are reasonable. 
Thus, we use semi-dense LiDAR depth as label, which is created by accumulating multiple LiDAR frames. 
With ego motion and calibration parameters, all static points can be transformed to destination image frame. 
Moving points are compensated by bounding box poses at each frame, which are estimated by interpolating bounding boxes provided by nuScenes in key frames.

\SubSubSection{Occlusion Removal via Flow Consistency}
\label{sec:flowconsistency}
When a foreground object occludes some of the accumulated LiDAR points, the resulting dense depth may include depth artifacts as the occluded pixels appear in gaps in the foreground object. KITTI~\cite{Uhrig2017THREEDV} takes advantages of the depth from stereo images to filter out such occluded points. 
As no stereo images are available in nuScenes, we propose detecting and removing occluded LiDAR points based on optical-scene flow consistency.

\begin{figure}[t!]
\captionsetup{font=small}
	\centering
	\scalebox{1.09}{
    \begin{tabular}{@{}c@{}c@{}c@{}}
	\includegraphics[width=1.17in]{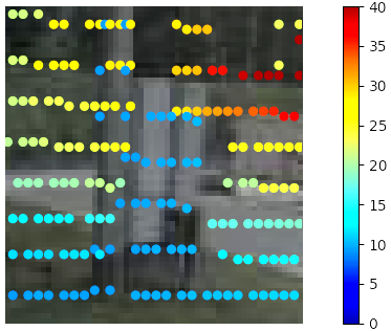} &
	\includegraphics[width=0.9in]{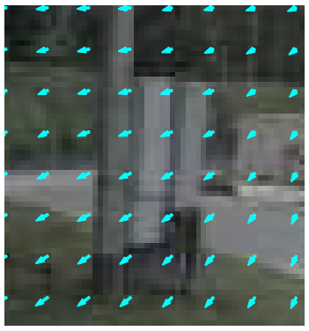}  &
	\includegraphics[width=0.9in]{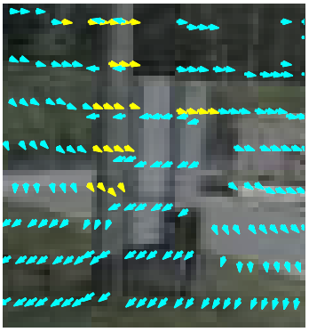} \vspace{-1mm}\\
     \footnotesize{ (a) LiDAR depth} &  \footnotesize{ (b) Optical flow} & \footnotesize{ (c) LiDAR flow} \vspace{-2mm}\\
    \end{tabular}   }
	\caption{\small An example of how LiDAR scene flow and optical flow differences are used to infer occlusions of LiDAR pixels. LiDAR flows are plotted as yellow if the $L_2$ norm of LiDAR/optical flow differences are larger than a threshold.  This is used in the accumulation of LiDAR for building ground-truth depth maps, see Fig.~\ref{Figure:flow_filter}. \vspace{-6mm}}
	\label{Figure:flows_lidar}
\end{figure}

\begin{figure}[t!]
\captionsetup{font=small}
	\centering
    \begin{tabular}{@{}c@{}c@{}}
		\includegraphics[trim=60 20 100 20,clip,width=1.53in]{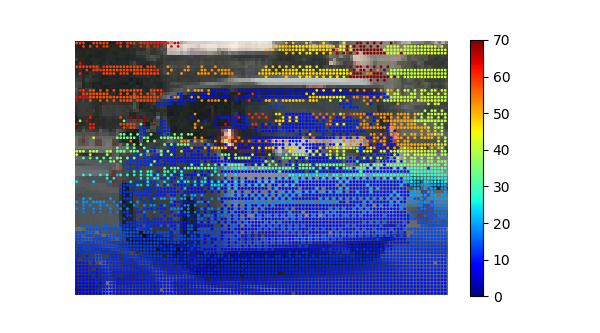} &
		\includegraphics[trim=60 20 60 20,clip,width=1.75in]{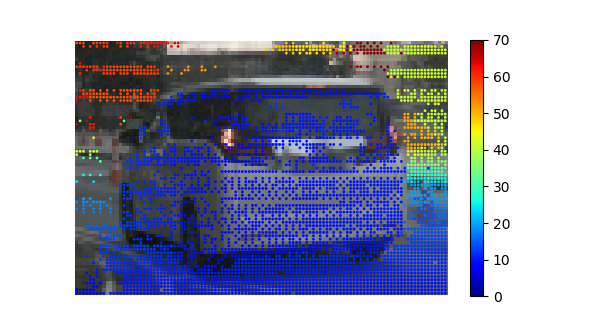} \vspace{-2mm} \\
    {\small (a) } & {\small (b) }
    \end{tabular}
    \vspace{-3mm}
	\caption{\small An example of using LiDAR flow and optical flow consistency to filter occluded pixels. (a) Accumulated LiDAR depth, and (b) Accumulated LiDAR depth with flow consistency filtering. \vspace{-6mm}}
	\label{Figure:flow_filter}
\end{figure}

The scene flow of LiDAR points, termed \emph{LiDAR flow}, is computed by projecting LiDAR points into two neighboring images and measuring the change in their coordinates.  
On moving objects, the point's positions  are corrected with the object motion.  
On static visible objects, LiDAR flow will equal optical flow, while on occluded surfaces LiDAR flow is usually different from the optical flow at the same pixel, see Fig.~\ref{Figure:flows_lidar}.
We calculate optical flow with~\cite{teed2020raft} pretrained on KITTI, and measure the difference between the two flows at the same pixel via the $L_2$ norm of their difference. 
Points with flow difference larger than a threshold $T_f$ are discarded as occluded points. Fig.~\ref{Figure:flow_filter} shows an example of using flow consistency to filter out occluded LiDAR depths.

\SubSubSection{Occlusion Removal via Segmentation}
Flow-based occluded pixel removal may fail in two cases.  
When there is little to no parallax, both optical and scene flow will be small, and their difference becomes not measurable.  
This occurs mostly at long range or along the motion direction.  
Further, LiDAR flow on moving objects can in some cases be identical to the occluded LiDAR flow behind it. 
In both of these cases flow consistency is insufficient to remove occluded pixels from the final depth estimate.

\begin{figure}[t!]
\captionsetup{font=small}
	\vspace{-1mm}
	\centering
	\scalebox{1}{
    \begin{tabular}{@{}c@{\hspace{0.3em}}c@{}}
	\includegraphics[trim=70 0 60 0,clip,width=1.6in]{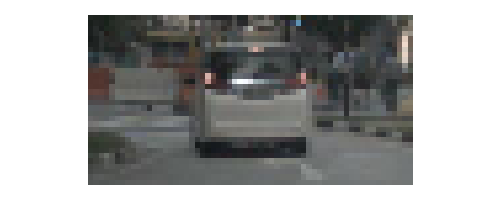} &
	\includegraphics[trim=70 0 60 0,clip,width=1.6in]{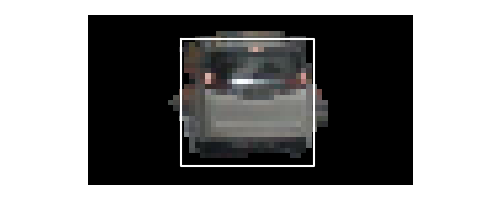} \vspace{-3mm}\\
     {\small (a) Car image} &  {\small (b) Semantic seg.~\& bound.~box}  \vspace{-1mm}\\
	\includegraphics[trim=70 0 35 0,clip,width=1.6in]{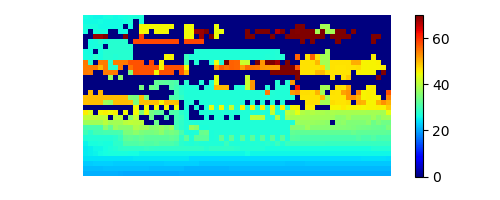} &
	\includegraphics[trim=70 0 35 0,clip,width=1.6in]{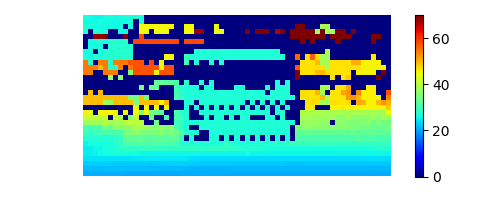}	 \vspace{-4mm}\\
    {\small (c) Depth before filtering} &  {\small (d) Depth after filtering} \vspace{-3mm}\\  
    \end{tabular}   }
	\caption{\small For small flow instances and some movers, flow consistency is insufficient to remove accumulated but occluded LiDAR pixels, see (a,c).  To remove these occluded pixels,  we first find vehicle pixels as the intersection between semantic segmentation and $2$D bounding box, see (b).  From the $3$D bounding box we know the maximum depth of the vehicle, and so can filter out all accumulated depths greater than this that are actually occluded, see (d). \vspace{-8mm}}
	\label{Figure:occlusion_box_seg}
\end{figure}

To solve this problem, we use a combination of $3$D bounding boxes and semantic segmentation to remove occluded points appearing on top of objects. First, accurate pixel region of an instance is determined by the intersection of $3$D bounding box projection and semantic segmentation.
The maximum depth of bounding box corners is used to decide whether LiDAR points falling on the object are on it or behind it. 
Points within the semantic segmentation and closer than this maximum distance are kept, while points in the segmentation and behind the bounding box are filtered out as occluded LiDAR points. Fig.~\ref{Figure:occlusion_box_seg} shows an example of removing occluded points appearing on vehicle instances. We use a semantic segmentation model~\cite{cheng2020panoptic} pre-trained with CityScape~\cite{cordts2016cityscapes} to segment vehicle pixels.

\begin{figure*}[t!]
\captionsetup{font=small}
	\centering
	\scalebox{0.7}
	{
    \begin{tabular}{@{}c@{}c@{}c@{}c@{}c@{}c@{}c@{}}
\includegraphics[width=1.3in]{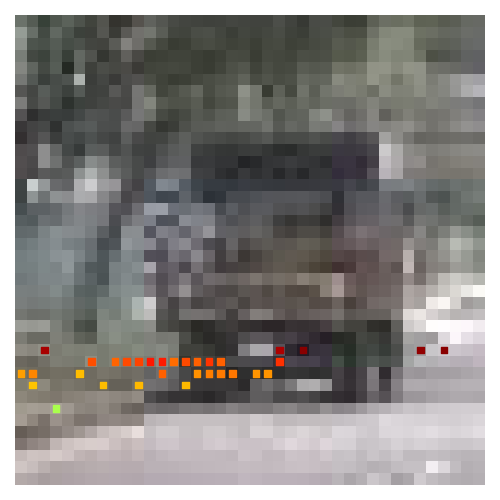} &
\includegraphics[width=1.3in]{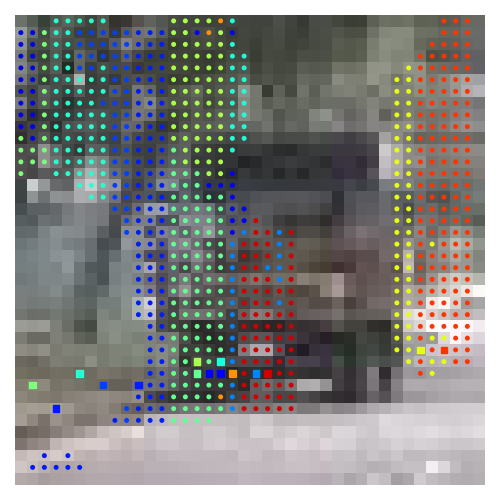} &
\includegraphics[width=1.3in]{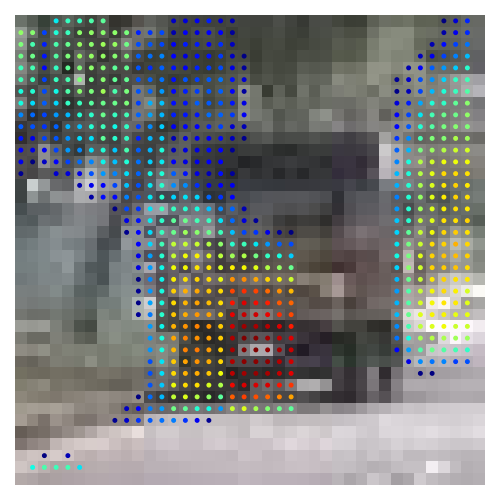} &
\includegraphics[width=1.3in]{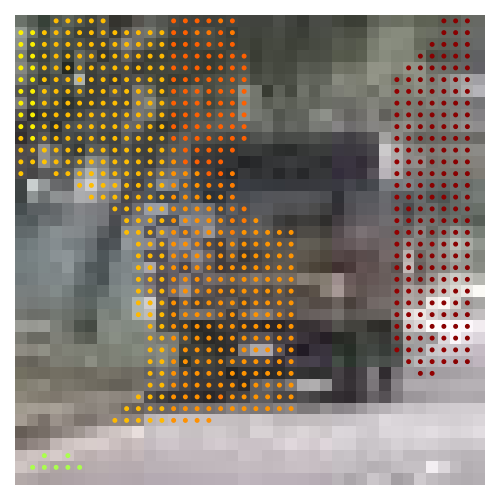} &
\includegraphics[width=1.3in]{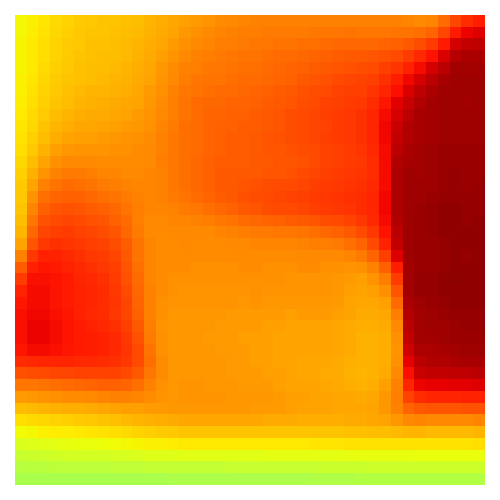} &
\includegraphics[width=1.3in]{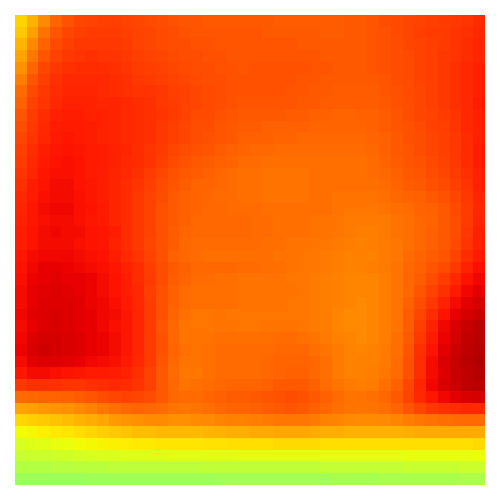} &
\includegraphics[width=1.3in]{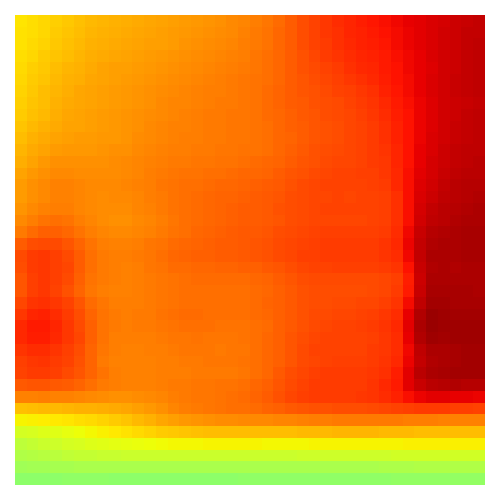} \vspace{-2mm}\\

\includegraphics[width=1.3in]{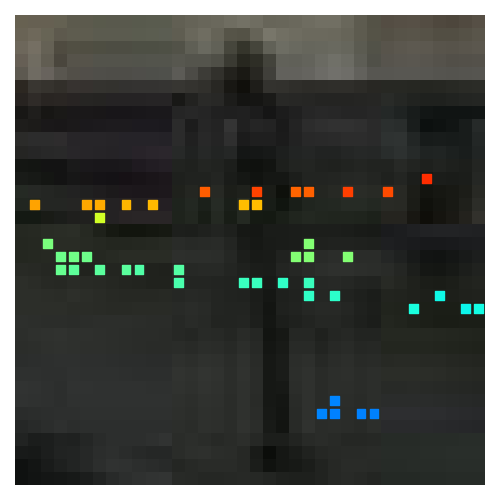} &
\includegraphics[width=1.3in]{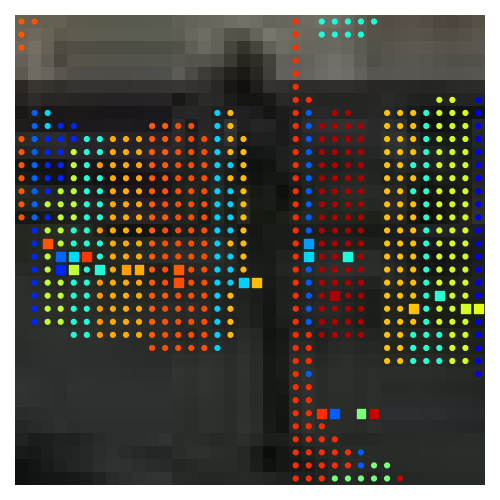} &
\includegraphics[width=1.3in]{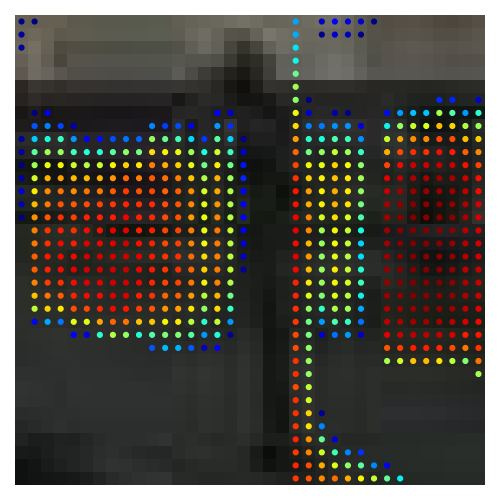} &
\includegraphics[width=1.3in]{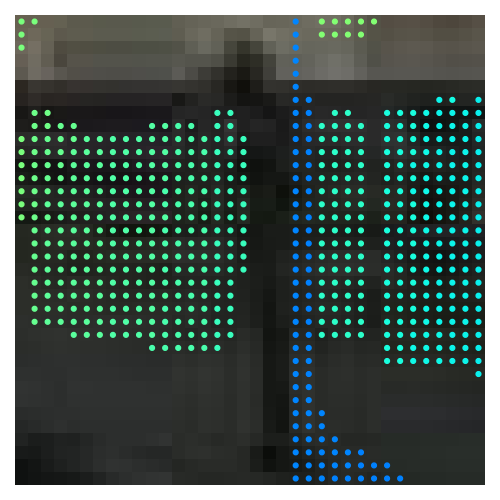} &
\includegraphics[width=1.3in]{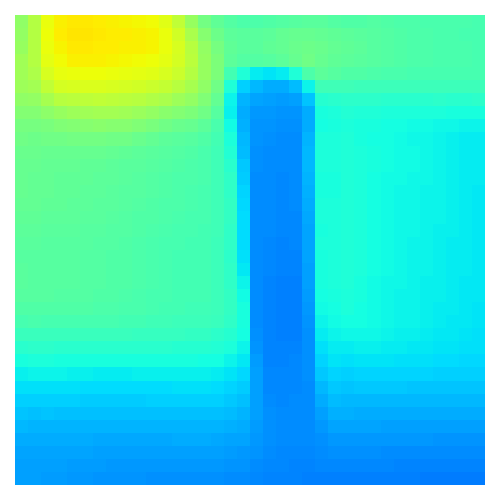} &
\includegraphics[width=1.3in]{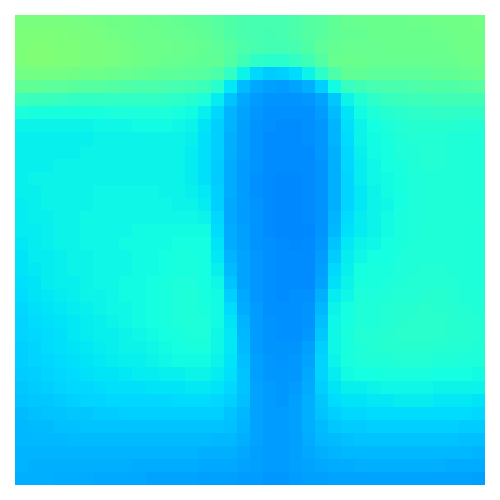} &
\includegraphics[width=1.3in]{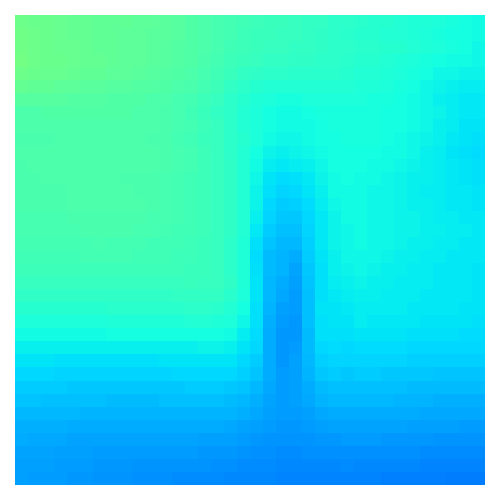} \vspace{-2mm}\\

\includegraphics[width=1.3in]{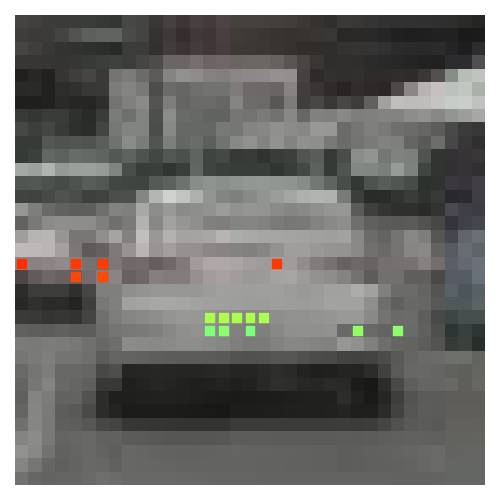} &
\includegraphics[ width=1.3in]{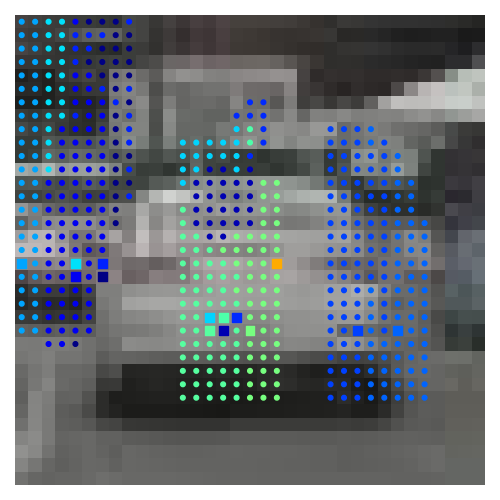} &
\includegraphics[width=1.3in]{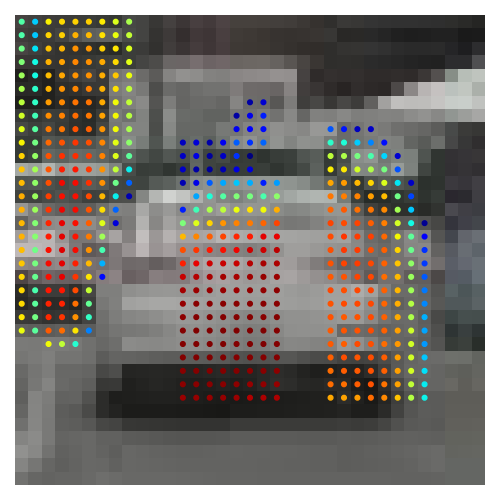} &
\includegraphics[width=1.3in]{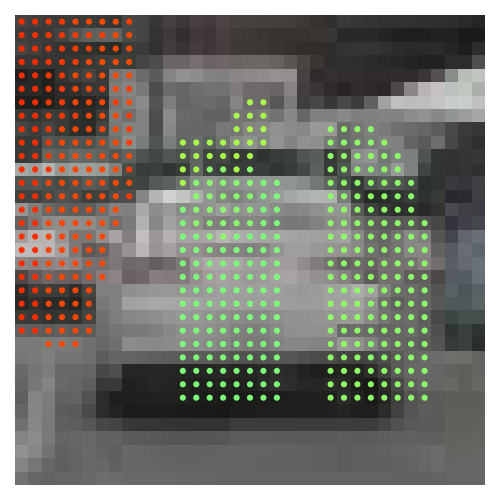} &
\includegraphics[width=1.3in]{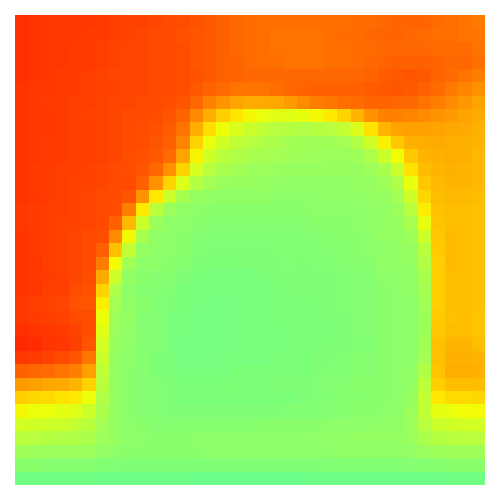} &
\includegraphics[width=1.3in]{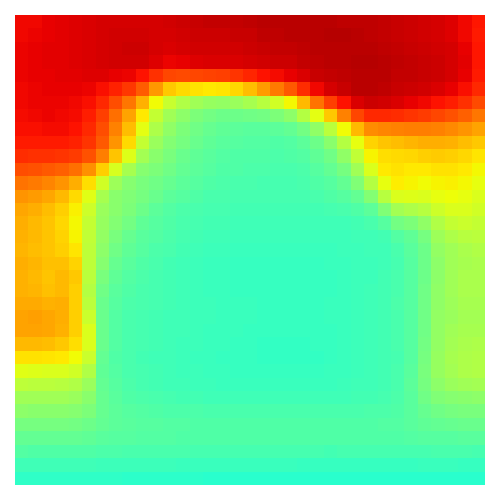} &
\includegraphics[width=1.3in]{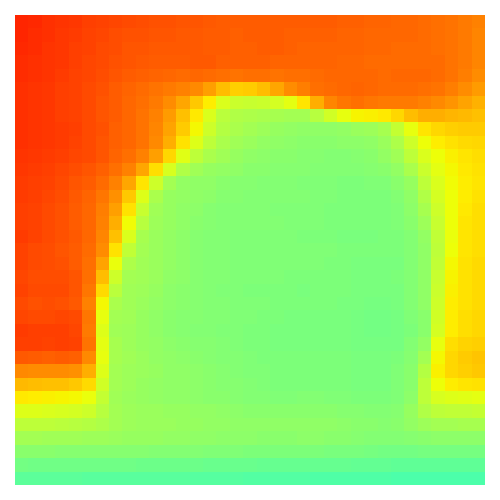} \vspace{-2mm}\\

\includegraphics[width=1.3in]{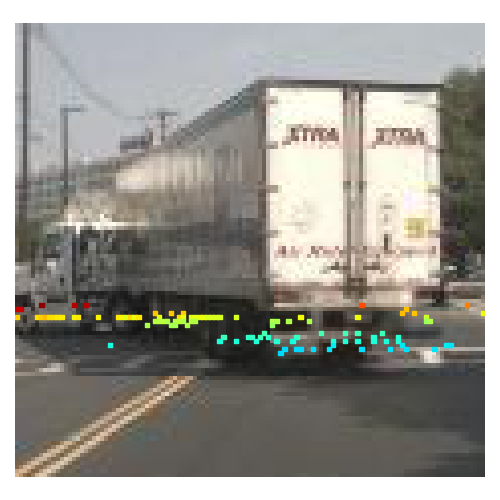} &
\includegraphics[width=1.3in]{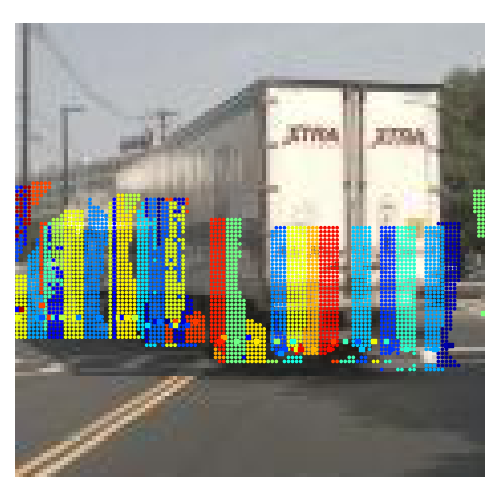} &
\includegraphics[width=1.3in]{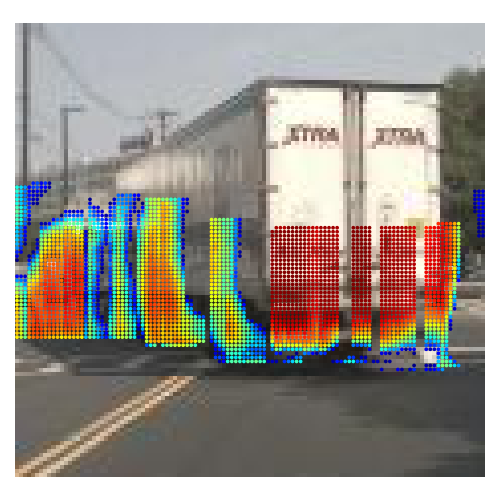} &
\includegraphics[width=1.3in]{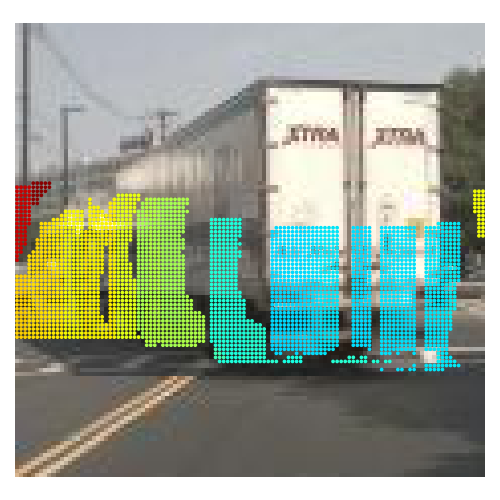} &
\includegraphics[width=1.3in]{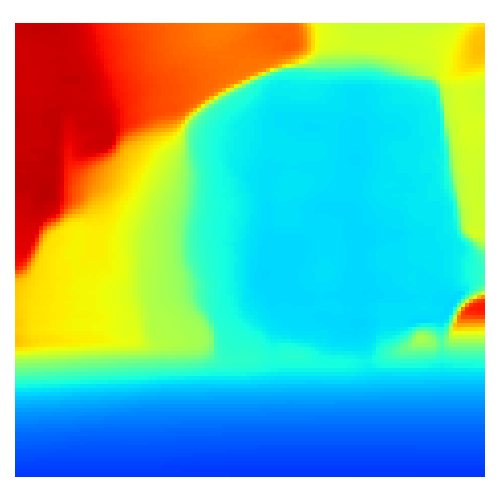} &
\includegraphics[width=1.3in]{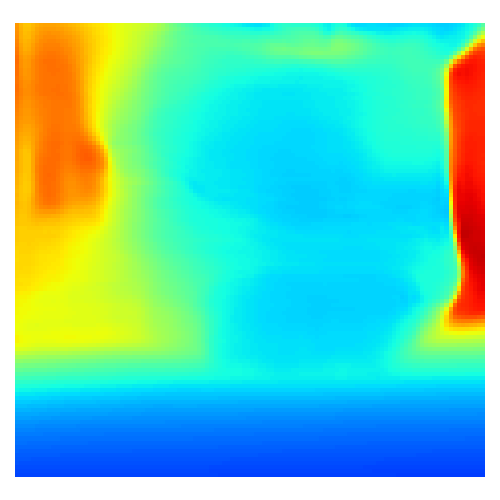} &
\includegraphics[width=1.3in]{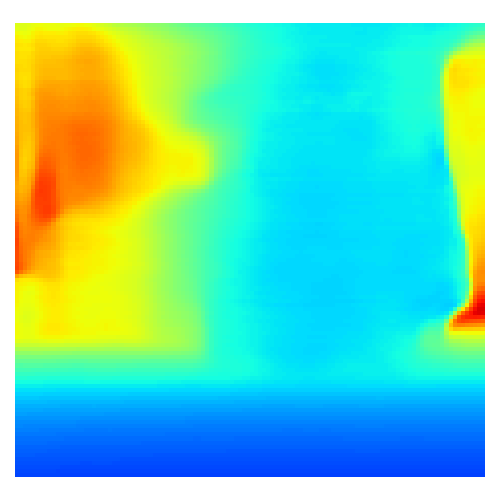} \vspace{-2mm}\\

\includegraphics[trim = 0 0 0 440, clip, width=1.3in]{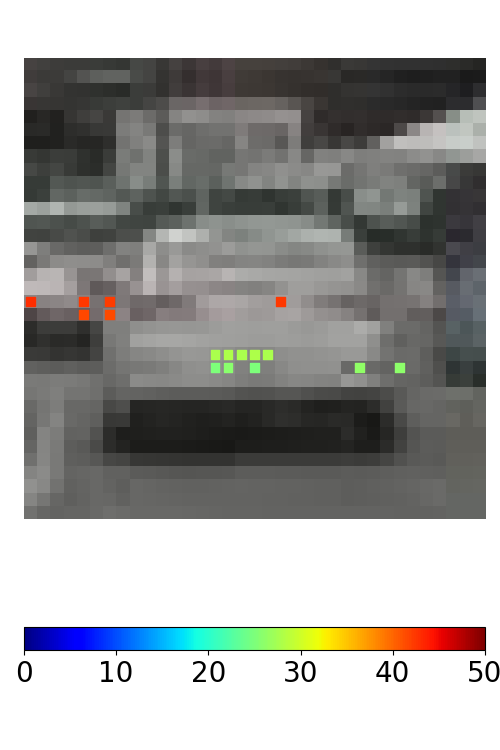} &
 &
\includegraphics[trim = 0 0 0 440, clip, width=1.3in]{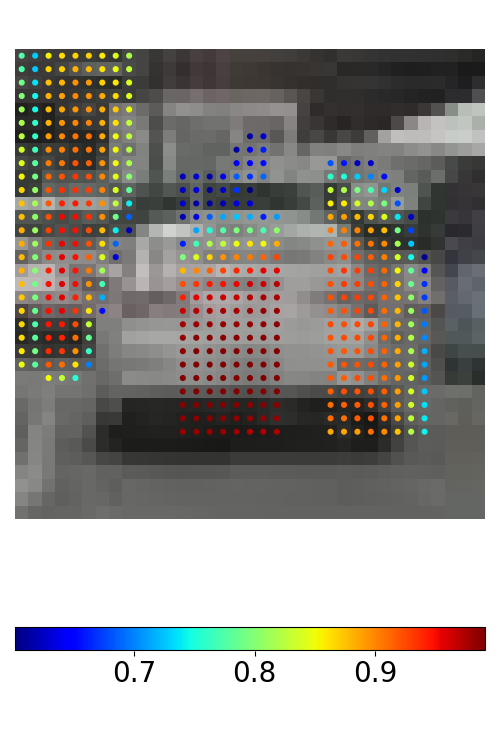} &
\includegraphics[trim = 0 0 0 440, clip, width=1.3in]{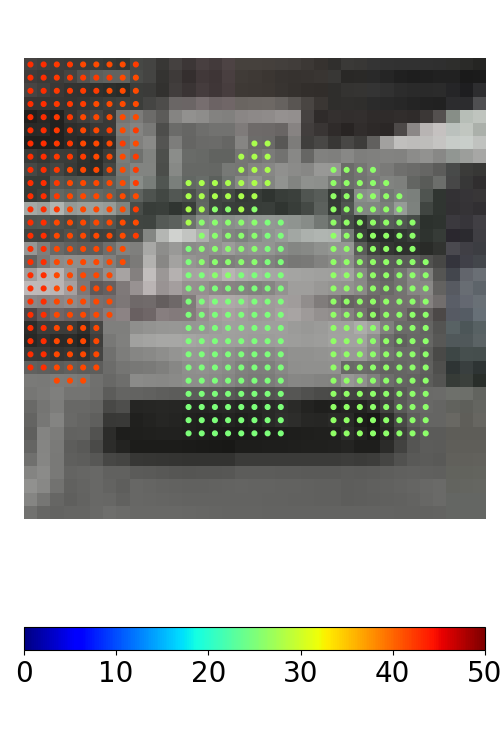} &
\includegraphics[trim = 0 0 0 440, clip, width=1.3in]{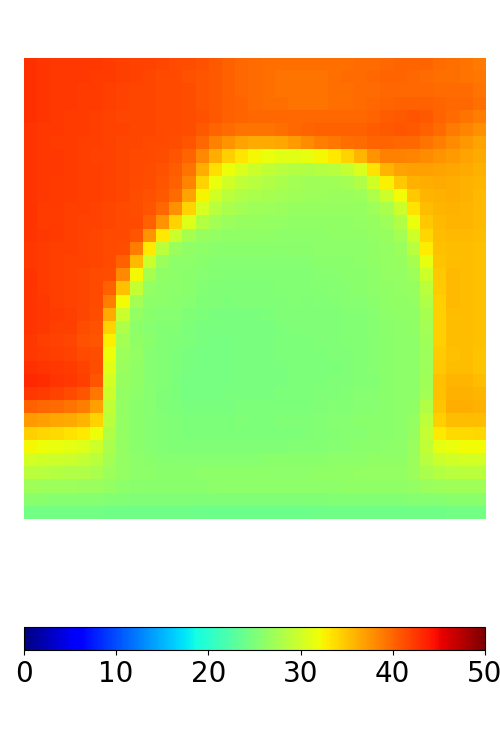} &
\includegraphics[trim = 0 0 0 440, clip, width=1.3in]{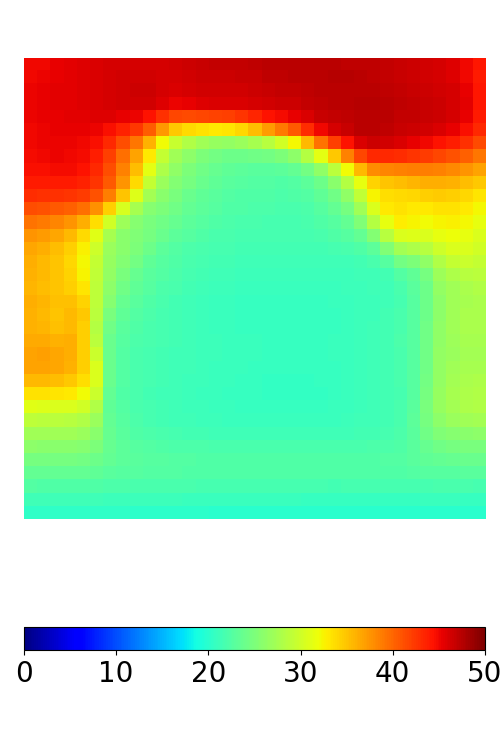} &
\includegraphics[trim = 0 0 0 440, clip, width=1.3in]{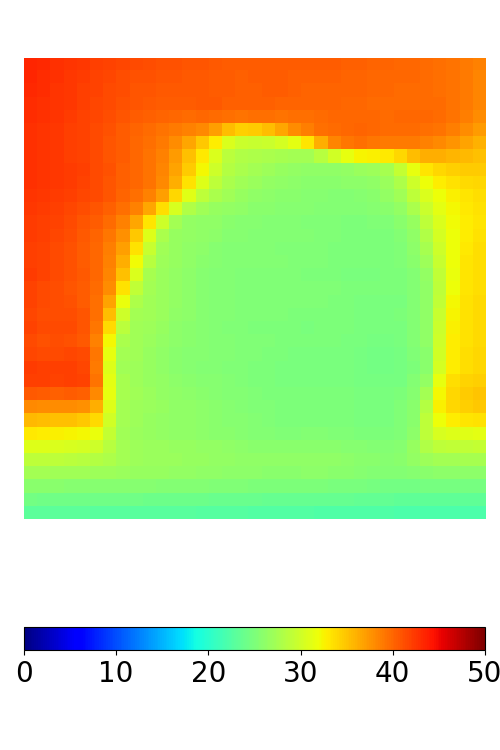} \vspace{-5mm}\\

    {\small (a) } & {\small (b) }  & {\small (c)}  & {\small (d)}  & {\small (e)}  & {\small (f)} & {\small (g)} \vspace{-3mm} \\
\end{tabular}}
	\caption{\small (a) Raw radar depths (b) Each color pixel with a maximum RC-PDA $>0.6$ is marked with a color indicating which radar pixel it is associated with. (c) The RC-PDA score with values $>0.6$ for each pixel. (d) The MER channel with RC-PDA $>0.6$. (e) Our final predicted depth. (f) Depth from monocular input to Stage $2$. (g) Depth from monocular and raw radar input to Stage $2$.}
	\label{Figure:vis_aff}
\end{figure*}

\SubSection{Algorithm Summary}

We propose a two-stage depth estimation process, as in Fig.~\ref{fig:network_arch}.  The Stage $1$ estimates RC-PDA for each radar pixel, which is transformed into our MER representation as detailed in Sec.~\ref{sec:pda_mer} and fed into Stage $2$ which performs conventional depth completion.  Both stages are supervised by the accumulated dense LiDAR, with pixels not having a LiDAR depth given zero weight.  
Network $1$ uses an encoder-decoder network with skip connections similar to U-Net~\cite{Ronneberger:2015:unet} and~\cite{ma2019self} with details in supplementary material.

\Section{Experimental Results}

\noindent\textbf{Dataset} We train and test on a subset of images from the nuScenes dataset~\cite{nuscenes2019}, including $12,610$, $1,628$ and $1,623$ samples for training, validation and testing, respectively. 
The data are collected with moving ego vehicle so flow calculation described in Sec.~\ref{sec:flowconsistency} can be applied. The depth range for training and testing is $0$-$50$ meters. Resolutions of inputs and outputs are $400\times192$.
As described in Sec.~\ref{sec:lidar-supervision}, we build semi-dense depth images by accumulating LiDAR pixels from $21$ subsequent frames and $4$ previous frames (sampled every other frame), and use these for supervision.

\noindent\textbf{Implementation details}
For parameters, we use $T_a=1$\,m and $T_r=0.05$ for Eq.~\ref{eq:aff_label}. In Sec.~\ref{sec:flowconsistency} we use $T_f=3$ to decide flow consistency. MER has $6$ channels with $T_1$ to $T_6$ set as $0.5$, $0.6$, $0.7$, $0.8$, $0.9$ and $0.95$, respectively. 
At Stage $1$, we use a U-Net with $5$ levels of resolutions and $180$ output channels, corresponding to $180$ pixels in a rectangle neighborhood with size $w=5$ and $h=36$. As the radar points are typically on the lower part of image, to fully leveraging the neighborhood, the neighborhood center is below the rectangle center with $30$ pixel above, $5$ pixels below and $2$ pixels on left and right to provide more space for radar points to extend upwards. At Stage $2$, we employ two existing depth completion architectures,~\cite{ma2018sparse} and~\cite{li2020multi}, originally designed for LiDAR-camera pairs.

\begin{figure*}[t!]
\captionsetup{font=small}
	\centering
	\scalebox{0.85}
	{
    \begin{tabular}{@{}c@{}c@{}c@{}c@{}}
\includegraphics[width=2in]{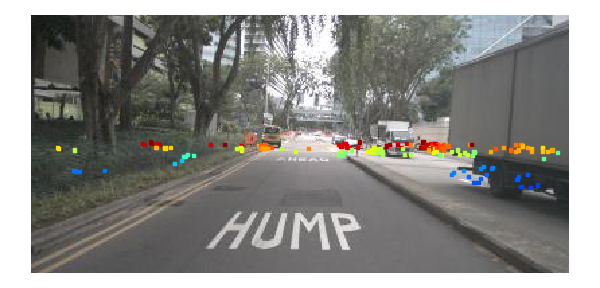}&
\includegraphics[width=2in]{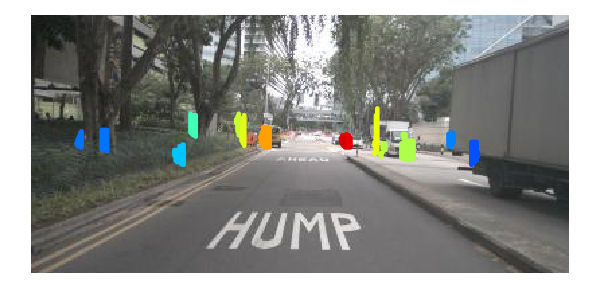}&
\includegraphics[width=2in]{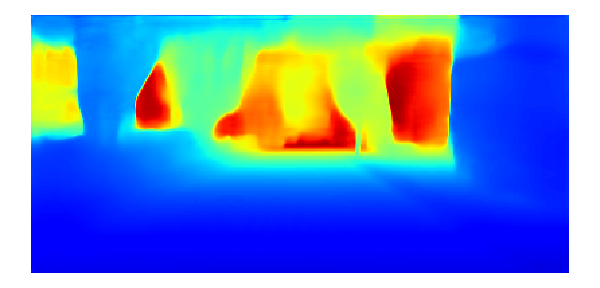}&
\includegraphics[trim=0cm 0.4cm 0cm 0cm, clip=true, width=2.35in]{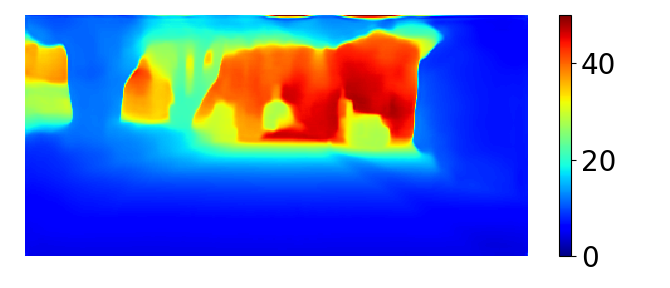} \vspace{-2.3mm}\\

\includegraphics[width=2in]{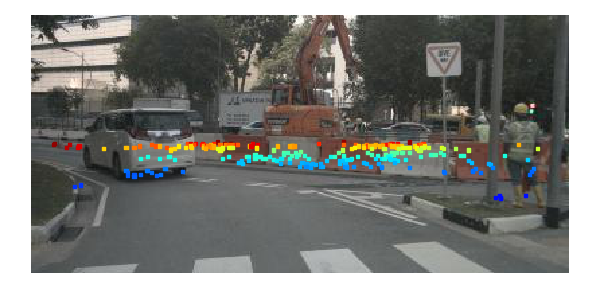}&
\includegraphics[width=2in]{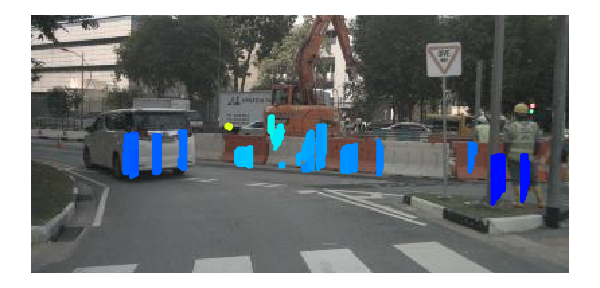}&
\includegraphics[width=2in]{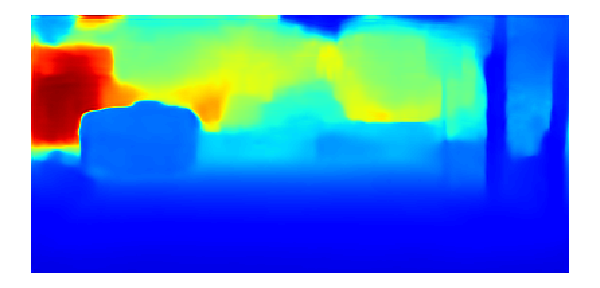}&
\includegraphics[trim=0cm 0.4cm 0cm 0.1cm, clip=true, width=2.35in]{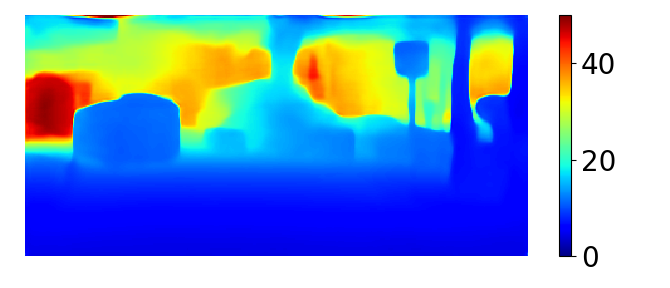} \vspace{-2mm}\\
{\small (a) } & {\small (b) }  & {\small (c)}  & {\small (d)}\\
    \end{tabular}}
    \vspace{-2.5mm}
    	\caption{\small Qualitative depth completion comparison showing gains from using MER over raw radar.  (a) Raw radar on top of image versus (b) A MER channel with RC-PDA $>0.8$ on top of image. Depth completion (c) without and (d) with using MER.\vspace{-2mm}}
	\label{Figure:vis_cmp}
\end{figure*}

\SubSection{Visualization of Predicted RC-PDA}
The predicted RC-PDA and estimated depths from Stage $2$~\cite{ma2018sparse} are visualized in Fig.~\ref{Figure:vis_aff}.  Column (a) shows the raw radar pixels plotted on images and often include occluded radar pixels.  Column (b) shows how image pixels are associated with different radar pixels according to their maximum RC-PDA. Radar pixels and their associated neighboring pixels are marked with the same color.  Notice in column (c) that RC-PDA is high within objects and decreases after crossing boundaries. Occluded radar depth are mostly discarded as their predicted RC-PDA is low. In column (e), the dense depths predicted from MER are improved over predictions from (g) raw radar and/or (f) monocular. For example, in Row $2$ of Fig.~\ref{Figure:vis_aff}, our predicted pole depth in (e) has better boundaries than monocular-only in (f), and monocular plus raw radar in (g).  
How we achieve this can be intuitively understood by comparing raw radar in (a) with our MER in (d), the output of Stage $1$.  While raw radar has many incorrect depths, MER selects correct radar depths and extends these depths along the pole and background, enabling improved final depth inference.

\begin{figure}[t!]
\captionsetup{font=small}
	\centering
	\begin{tabular}{c}
	\includegraphics[width=2.8in]{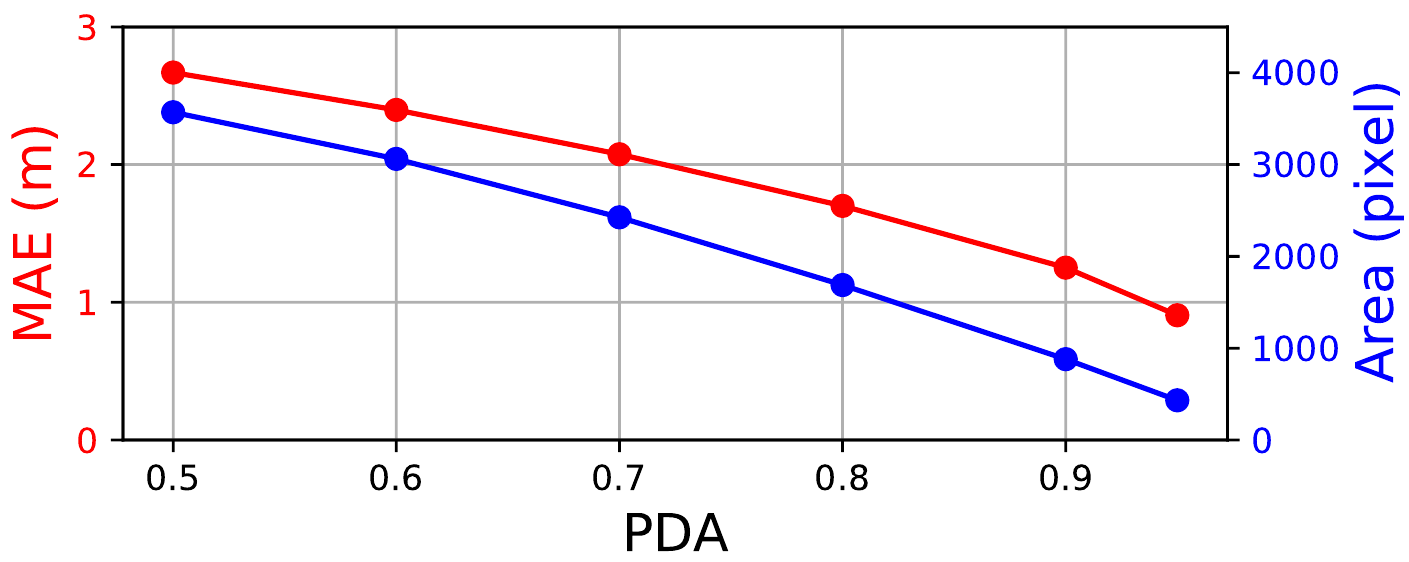} \\
	\end{tabular}
	\vspace{-3mm}
	\caption{\small Image area and depth error of enhanced radar in MER for regions with minimal RC-PDA at $6$ confidence levels. \vspace{-3mm}}
	\label{Figure:aff_curve}
\end{figure}

\SubSection{Accuracy of MER}
To be useful in improving radar-camera depth completion, the enhanced radar depth in the vicinity of radar points should be better than alternatives. We compare the depth error of the MER in regions where RC-PDA is $>0.9$, with a few baseline methods and results are shown in Tab.~\ref{tab:expand_error}. 
The enhanced radar depth from Stage $1$ improves over not only raw radar depth but also depth estimates from Stage $2$ using monocular as well as monocular plus radar. In comparison, Stage $2$ keeps the accuracy of the enhanced radar depth when using it for depth completion. The depth error for raw radar depth is very large since many of them are occluded and far behind foreground. About $35\%$ of radar points in the test frames have a maximum RC-PDA smaller than $T_1$ in their neighborhood and are discarded as occluded points. 
Further, Fig.~\ref{Figure:aff_curve} shows the depth error and per-image average area  of expanded depth from $6$ MER channels, respectively. 
It shows, as confidence increases, higher RC-PDA corresponds to higher accuracy and smaller expanded areas.

\begin{table}[t]
\captionsetup{font=small}
	\begin{center}	
		\scalebox{0.7}{
			\begin{tabular}{|c|c|cccc|}
				\hline
				Network $2$				& Input &  MAE & Abs Rel & RMSE & RMSE log\\ 
				\hline
				& Image               & $2.385$ & $0.110$ & $3.505$ & $0.150$ \\
				Ma~\emph{et al.}~\cite{ma2018sparse} 
				& Image, radar        & $1.609$ & $0.078$ & $2.865$ & $0.126$ \\				
				& Image, radar, MER   & $\mathbf{1.229}$ & $\mathbf{0.058}$ & $\mathbf{2.651}$ & $\mathbf{0.114}$ \\
				\hline
				\hline
				\multirow{2}{*}{Li~\emph{et al.}~\cite{li2020multi}} 
				& Image, radar        & $1.759$ & $0.084$ & $3.039$ & $0.133$ \\
				& Image, radar, MER   & $\mathbf{1.274}$ & $\mathbf{0.061}$ & $\mathbf{2.670}$ & $\mathbf{0.116}$ \\
				\hline
				\hline
				\multirow{2}{*}{None} 
				& MER   & $1.251$ & $0.059$ & $2.701$ & $0.117$ \\
				& Radar & $7.369$ & $0.475$ & $10.900$ & $0.448$ \\
				\hline
		\end{tabular} }
	\end{center}
	\vspace{-5mm}
	\caption{\small Depth error (m) in image regions around non-occluded radar returns, defined as regions with RC-PDA $>0.9$.\vspace{-3mm}}
	\label{tab:expand_error}
	\vspace{-0.5mm}
\end{table}

\SubSection{Comparison of Depth Completion}
To evaluate effectiveness of the enhanced radar depth in depth completion, we compare the depth error with and without using MER as input for Network $2$, and show performance in Tabs.~\ref{tab:d_error} and~\ref{tab:d_error_low}. The results show that including radar improves depth completion over monocular, while using our proposed MER further improves the accuracy of depth completion for the same network.

Qualitative comparisons between depth completion~\cite{ma2018sparse} with and without using MER are shown in Fig.~\ref{Figure:vis_cmp}.  
This shows improvement from MER in estimating object depth boundaries including close objects (such as the traffic sign on the bottom image) and far objects.

\begin{table}[t]
\captionsetup{font=small}
	\begin{center}	
		\scalebox{0.7}{
			\begin{tabular}{|c|c|cccc|}
				\hline
				Network $2$ 				& Input &  MAE & Abs Rel & RMSE & RMSE log\\ 
				\hline
				& Image               & $1.808$ & $0.102$ & $3.552$ & $0.160$ \\
				Ma~\emph{et al.}~\cite{ma2018sparse} 
				& Image, radar        & $1.569$ & $0.090$ & $3.327$ & $0.152$ \\				
				& Image, radar, MER   & $\mathbf{1.472}$ & $\mathbf{0.085}$ & $\mathbf{3.179}$ & $\mathbf{0.144}$ \\
				\hline
				\hline
				\multirow{2}{*}{Li~\emph{et al.}~\cite{li2020multi}} 
				& Image, radar        & $1.821$ & $0.107$ & $3.650$ & $0.170$ \\
				& Image, radar, MER   & $\mathbf{1.655}$ & $\mathbf{0.094}$ & $\mathbf{3.463}$ & $\mathbf{0.159}$ \\
				\hline
		\end{tabular} }
	\end{center}
	\vspace{-5mm}
	\caption{\small Full-image depth estimation/completion errors (m).\vspace{-2mm}}
	\label{tab:d_error}
	
\end{table}

\begin{table}[t]
\captionsetup{font=small}
	\begin{center}	
		\scalebox{0.7}{
			\begin{tabular}{|c|c|cccc|}
				\hline
				Network $2$ 				& Input &  MAE & Abs Rel & RMSE & RMSE log\\ 
				\hline
				& Image               & $2.673$ & $0.153$ & $4.259$ & $0.202$ \\
				Ma~\emph{et al.}~\cite{ma2018sparse} 
				& Image, radar        & $2.263$ & $0.134$ & $4.028$ & $0.194$ \\				
				& Image, radar, MER   & $\mathbf{2.078}$ & $\mathbf{0.124}$ & $\mathbf{3.864}$ & $\mathbf{0.183}$ \\
				\hline
				\hline
				\multirow{2}{*}{Li~\emph{et al.}~\cite{li2020multi}} 
				& Image, radar        & $2.515$ & $0.154$ & $4.266$ & $0.211$ \\	
				& Image, radar, MER   & $\mathbf{2.189}$ & $\mathbf{0.132}$ & $\mathbf{3.943}$ & $\mathbf{0.193}$\\		 
				\hline
		\end{tabular} }
	\end{center}
	\vspace{-5mm}
	\caption{\small Depth  estimation/completion errors (m) in the low-height region ($0.3$-$2$ meters above ground).\vspace{-3mm}}
	\label{tab:d_error_low}
	
\end{table}

\Section{Conclusion}
Radar-based depth completion introduces additional challenges and complexities beyond LiDAR-based depth completion.  A significant difficulty is the large ambiguity in associating radar pixels with image pixels.  We address this with RC-PDA, a learned measure that associates radar hits with nearby image pixels at the same depth.  From RC-PDA we create an enhanced and densified radar image called MER.  Our experiments show that depth completion using MER achieves improved accuracy over depth completion with raw radar.  As part of this work we also create a semi-dense accumulated LiDAR depth dataset for training depth completion on nuScenes. 

\vspace{2mm}
\noindent\textbf{Acknowledgement}
This work was supported by the Ford-MSU Alliance.

{\small
\bibliographystyle{ieee_fullname}
\bibliography{bib_abbrev, bib_strings, bib_lidarcam, bib_radarcam}
}

\end{document}